# Non-invasive liver fibrosis screening on CT images using radiomics

Jay J. Yoo[a,b,f], Khashayar Namdar[a,b,f], Sean Carey[g], Sandra E. Fischer[m], Chris McIntosh[f,g,i,j,k,l], Farzad Khalvati[a,b,c,d,e,f,*], Patrik Rogalla[g,l]

[a] *Institute of Medical Science, University of Toronto, Toronto M5S 1A8, CA*

[b] *Department of Diagnostic Imaging, Research Institute, The Hospital for Sick Children, Toronto M5G 1X8, CA*

[c] *Department of Medical Imaging, University of Toronto, Toronto M5S 1A8, CA*

[d] *Department of Computer Science, University of Toronto, Toronto M5S 1A8, CA*

[e] *Department of Mechanical and Industrial Engineering, University of Toronto, Toronto M5S 1A8, CA*

[f] *Vector Institute, Toronto M5G 1M1, CA*

[g] *Joint Department of Medical Imaging, University Health Network, Toronto M5G 2C4, CA*

[h] *Department of Radiology, Princess Margaret Cancer Centre, Toronto M5G 2C1, CA*

[i] *Department of Medical Biophysics, University of Toronto, Toronto M5S 1A8, CA*

[j] *Techna Institute, University Health Network, Toronto M5G 2C4, CA*

[k] *Radiation Medicine Program, Princess Margaret Cancer Centre, Toronto M5G 2C1, CA*

[l] *Peter Munk Cardiac Center, University Health Network, Toronto M5G 2N2, CA*

[m] *Department of Pathology, University Health Network, Toronto M5G 2C4, CA*

[*] *Corresponding Author, E-mail Address: farzad.khalvati@utoronto.ca, Telephone Number: 416-813-7654 X 309305*

*Farzad Khalvati and Patrik Rogalla are co-senior authors.*

## Abstract

**Objectives:** To develop and evaluate a radiomics machine learning model for detecting liver fibrosis on CT of the liver.

**Methods:** For this retrospective, single-centre study, radiomic features were extracted from Regions of Interest (ROIs) on CT images of patients who underwent simultaneous liver biopsy and CT examinations. Combinations of contrast, normalization, machine learning model, and feature selection method were determined based on their mean test Area Under the Receiver Operating Characteristic curve (AUC) on randomly placed ROIs. The combination and selected features with the highest AUC were used to develop a final liver fibrosis screening model.

**Results:** The study included 101 male and 68 female patients (mean age = 51.2 years ± 14.7 [SD]). When averaging the AUC across all combinations, non-contrast enhanced (NC) CT (AUC, 0.6100; 95% CI: 0.5897, 0.6303) outperformed contrast-enhanced CT (AUC, 0.5680; 95% CI: 0.5471, 0.5890). The combination of hyperparameters and features that yielded the highest AUC was a logistic regression model with inputs features of maximum, energy, kurtosis, skewness, and small area high gray level emphasis extracted from non-contrast enhanced NC CT normalized using Gamma correction with γ = 1.5 (AUC, 0.7833; 95% CI: 0.7821, 0.7845), (sensitivity, 0.9091; 95% CI: 0.9091, 0.9091).

**Conclusions:** Radiomics-based machine learning models allow for the detection of liver fibrosis with reasonable accuracy and high sensitivity on NC CT. Thus, these models can be used to non-invasively screen for liver fibrosis, contributing to earlier detection of the disease at a potentially curable stage.

**Clinical relevance statement:** Non-invasive liver fibrosis detection on CT images can be seamlessly incorporated into liver protocols, enabling screenings for individuals undergoing CT imaging for any clinical reason. This may improve the frequency of early detection of liver fibrosis, therefore supporting early interventions.

# Keywords

Machine Learning; Liver; Fibrosis; Humans; Tomography; X-Ray Computed

# Key points

- Machine learning models trained on radiomic features were used to enable non-invasive liver fibrosis detection on CT images.
- Logistic regression models trained on non-contrast enhanced images normalized using Gamma normalization (γ = 1.5) with maximum, energy, kurtosis, skewness, and small area high gray level emphasis performed best for fibrosis detection on CT images across all evaluated hyperparameter combinations.
- The presented model can be used to screen for liver fibrosis on existing CT without additional acquisitions, improving likelihood of detecting fibrosis early, when it is most treatable.

# Abbreviations and acronyms

- **3D:** 3-dimensional
- **AUC:** Area under the receiver operating characteristic curve
- **CE:** Contrast-enhanced
- **CT:** Computed tomography
- **Gamma-0.5:** Gamma correction with γ = 0.5
- **Gamma-1.5:** Gamma correction with γ = 1.5
- **HU:** Hounsfield unit
- **LASSO:** Least absolute shrinkage and selection operator
- **ML:** Machine learning
- **MRI:** Magnetic resonance imaging
- **NASH:** Nonalcoholic steatohepatitis
- **NC:** Non-contrast-enhanced

- **PCA:** Principal component analysis
- **RBF:** Radial basis function
- **ROI:** Region of interest
- **SAG:** Stochastic average gradient
- **SAGA:** Stochastic average gradient ascent
- **SGD:** Stochastic gradient descent
- **SMOTE:** Synthetic minority oversampling technique
- **SVM:** Support vector machine
- **US:** Ultrasound

## Introduction

Fibrosis is the wound-healing response to liver injury that can lead to cirrhosis and is associated with an increase in liver-related complications in patients with non-alcoholic steatohepatitis (NASH) [1]. Accurate diagnosis in the pre-cirrhotic phase enables early application of preventative and corrective measures to combat the progression of the disease. Therefore, it is desirable to screen the general population for liver NASH-fibrosis since individuals in the early stages of fibrosis are frequently asymptomatic, may exhibit normal transaminase blood levels, and may experience a deceleration in fibrosis progression during this phase. Image-guided liver biopsy is the diagnostic reference standard but is costly, can introduce risk of complications to patients, and demonstrates low intra- and interobserver repeatability [2][3]. Furthermore, due to the inconsistent geographic distribution of hepatic fibrosis, its accuracy suffers from a sampling error ranging between 55% and 75% [4]. These issues have led to an interest in developing non-invasive modalities for fibrosis detection [5][6].

The motivation to explore computed tomography (CT) over other imaging modalities is the possibility of seamlessly incorporating the required measurements into liver imaging protocols, which can potentially obviate the need for additional dedicated imaging acquisitions, such as magnetic resonance imaging (MRI) and ultrasound (US) elastography [10]. Hence, the detection of NASH-fibrosis on CT images performed for clinical indications other than liver diseases will enable the detection of fibrosis at early, subclinical stages and may therefore lead to early therapeutic interventions, ranging from simple lifestyle changes to novel treatment options.

Liver fibrosis is commonly staged into the 5-point METAVIR scoring system; F0: no fibrosis, F1: mild fibrosis, F2: moderate fibrosis, F3: severe fibrosis, F4: cirrhosis [11]. Prior studies have extensively evaluated the value of CT in detecting fibrosis stages of F2 and above [12][13][14]. However, differentiating healthy (F0) from diseased liver (F1-F4) on CT images is rarely explored. We hypothesize that a radiomics-based machine learning model trained on liver CT images (non-contrast and post-contrast) with biopsy proof will allow for differentiation of healthy livers from fibrotic livers of any stage.

# Materials and methods

## Data

This retrospective study was approved by the institutional Research Ethics Board (REB), and written informed consent was obtained from all patients.

The CT scans used in this study were acquired from a clinical trial that collected images spatially and temporally synchronized with percutaneous liver biopsies in patients with suspected diffuse liver disease from October 2019 to April 2021. Accepted indications were to rule out fibrosis due to autoimmune hepatitis (AIH), hepatitis C, chronic ETOH intake, and elevated enzymes of unknown clinical cause. Recruited patients underwent clinically indicated random liver biopsy, which is the accepted clinical reference standard for staging liver fibrosis. Ultrasound was used to guide the biopsy needle to a location in the right liver lobe, and then CT images were acquired (Aquilion ONE Genesis, Canon Medical Systems, Otawara, Japan) before (NC) and 70 seconds after injection of intravenous contrast material (CE) during a single breath-hold. Thin slice axial images were reconstructed from the projection data and resampled to $0.5 \times 0.5 \times 0.5$ mm voxel size to standardize the voxel sizes between images. 202 patients met the inclusion criteria of having clinical indication for random liver biopsy, 2 of which declined consent, resulting in 200 initial consecutive patients. The biopsy was clinically reported by subspecialized pathologists and then used to score each patient's CT on a 5-point scale from F0 to F4. Patient biopsies that could not be confidently scored were excluded. Patients with invalid metadata were also excluded. 15 patients did not have images due to ultrasound not being available to guide the biopsy needle, and images from 2 patients were not included due to the corresponding metadata being invalid. This led to 169 patients being included in this study. Seven patients did not have CE CT acquired because they did not consent to the use of contrast material, and one patient only had a CE CT due to the NC CT becoming corrupted, resulting in 168 NC CTs and 162 CE CTs from 169 patients. A flowchart of patient eligibility is presented in Figure 1.

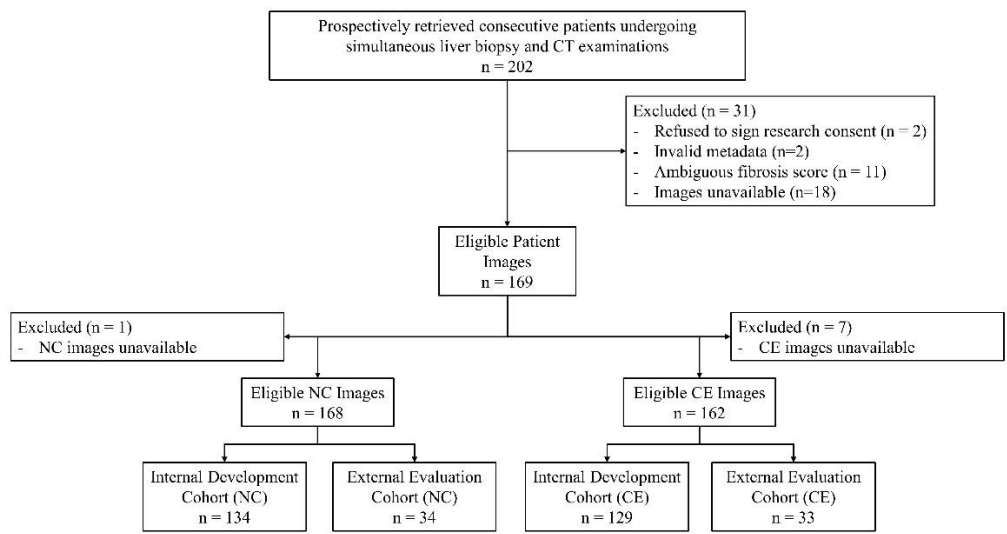

Figure 1: Flowchart of patient and image eligibility for this study.

The fibrosis stages were binarized into F0 and F1-F4, and these binary labels were used as the reference standard for this study. To reduce the impact of irrelevant information and prevent normalization methods from reducing the relevant voxels to insignificant attenuation values, the attenuation values for all CTs were thresholded as previously proposed [15]. NC CTs were thresholded to a range of 0 HU to 100 HU, and the attenuation values for CE CTs were thresholded to a range of -10 HU to 200 HU.

**Radiomics extraction**

We used PyRadiomics [16] to extract 1725 radiomic features (including first-order and second-order features) from biopsy-based and non-biopsy regions of interest (ROI) of 1.5 cm radius defined in each CT by one human reader. Shape features were excluded because they lack differentiative power between patients due to the ROIs being the same shape across all patients.

The first set of ROIs was placed such that their center was 2.5 cm distal to the biopsy needle in the right liver lobe to ensure that this set of ROIs was consistent with the corresponding area of biopsy. This set of ROIs is referred to as biopsy-based ROIs and served to eliminate a potential sampling error associated with the heterogeneous distribution of fibrosis in the liver.

Another set of ROIs was placed a minimum of 3 cm away from the centers of the biopsy-based ROIs. These ROIs were placed manually in a randomly chosen location with preference in the left liver lobe. This set of ROIs is referred to as non-biopsy ROIs and served to evaluate the ability of radiomics models to detect fibrosis in liver regions outside the biopsy-based ROIs.

Any ROIs not fully enclosed within the liver were shifted such that they were fully encapsulated within the liver parenchyma. Each shift was a maximum of 3 cm along any axis. An example ROI is presented in Figure 2, with 8 cm coverage in the coronal and sagittal views. The final position of the ROIs was verified by a subspecialized abdominal radiologist with 32 years of clinical experience.

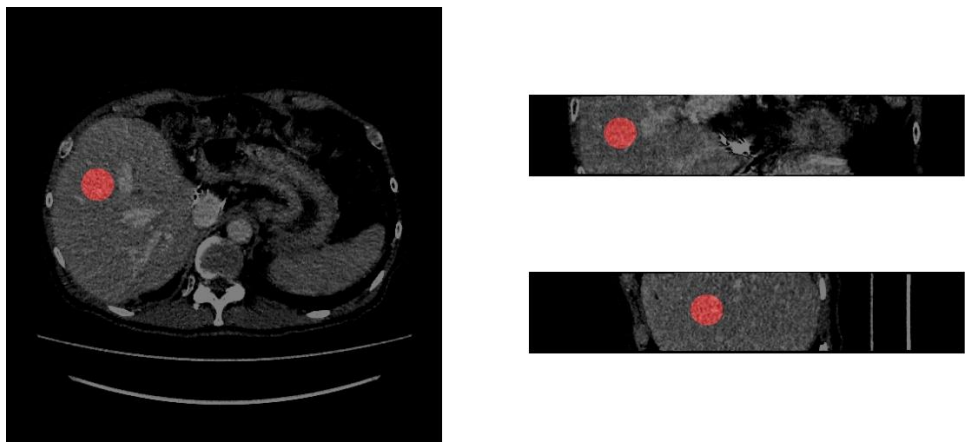

Figure 2: Example ROI (left: axial view, upper right: coronal view, lower right: sagittal view).

## Settings

We tested the following settings; Contrast: [NC, CE], Normalization: [None, Histogram Equalization, Min-max, Z-score, Gamma correction with γ = 0.5 (Gamma-0.5), Gamma correction with γ = 1.5 (Gamma-1.5)], Machine learning model type: [Logistic regression classifier, Random forest classifier, Support vector machines (SVM), Linear models with stochastic gradient descent (SGD) training], and Feature selection method: [None, Principal component analysis (PCA), Boruta, least absolute shrinkage and selection operator (LASSO) regression]. We evaluated every combination of these settings, which we refer to as configurations, totalling 1728 configurations.

## Experimental design

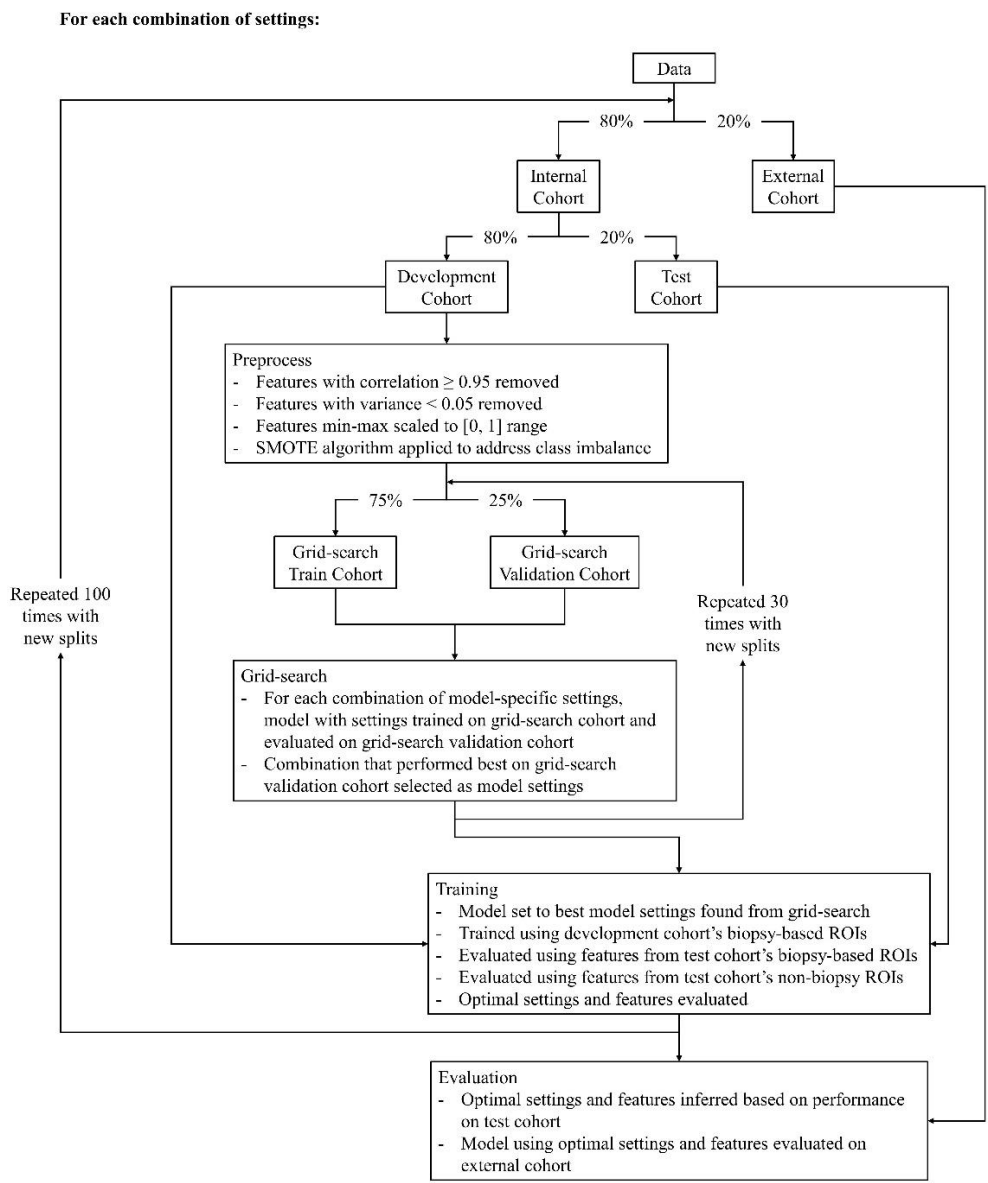

Figure 3: Flowchart of experimental design. Percentages in arrows represent percentage of data in each data split.

Figure 3 presents a flowchart of the experimentation design. We first split the NC and CE CTs into an internal cohort to be used for the experiments and an external held-out test cohort to be used for final evaluations using an 80/20 split. Inspired by Liu et al. [17] and Namdar et al. [18], 100 experiments were run for each configuration. For each experiment, the internal cohort was randomly split into a development cohort and an external cohort using 80/20 splits to increase confidence that the observed performance generalized to unseen data. Once split into the development and test cohorts, the cohorts were individually processed. First, redundant features with a correlation greater than 0.95 with another feature were removed. Then features with a variance of less than 0.05 were removed. Finally, the features in each cohort were min-max normalized to range [0, 1].

During each of the 100 experiments, the development set was further split into two sub-cohorts using 75/25 splits, referred to as grid-search train and grid-search validation, respectively. These sub-cohorts were used to perform a grid search of ML model-specific hyperparameters.

The internal cohort included 81 NC CTs and 76 CE CTs with fibrosis, as well as 53 CE CTs and 53 CE CTs without fibrosis. The external cohort included 23 NC CTs and 24 CE CTs with fibrosis, as well as 11 CE CTs and 9 CE CTs without fibrosis. Synthetic Minority Oversampling Technique (SMOTE) algorithm [19] was applied to the training data to address the class imbalances. The SMOTE algorithm synthesizes new data from the minority class by interpolating new data points from examples close to each other in the feature space of the minority class.

The grid search of ML model-specific hyperparameters was done by training ML models using each combination of model-specific hyperparameters 30 times using the grid-search train cohort and evaluating the area under the receiver operating characteristic curve (AUC) on the grid-search validation cohort across the 30 runs. The model-specific hyperparameters that resulted in the greatest mean AUC was applied to the model for the current configuration of settings. These ML model-specific settings are based on their implementations in the scikit-learn Python library and are presented in the supplementary material.

Once the model-specific hyperparameters were selected, they were used with the current configuration to train the model using the development cohort and were subsequently evaluated on the test cohort of patient CTs. The model was evaluated on both the features from the biopsy-based ROIs and the features from the non-biopsy ROIs of the test cohort, forming two sets of test results. The features from the non-biopsy ROIs were only used for evaluation and were not used to train the models. In clinical settings, the models would be applied to multiple randomly selected non-biopsy ROIs. The average prediction across the ROIs would then determine the final prediction.

The five most common features among the 100 experiments corresponding to the top five configurations, according to AUC, for the biopsy-based ROIs and the non-biopsy ROIs were used to train models that we will refer to as simple due to the low number of input features. The hyperparameters used for the simple models were the hyperparameters that yielded the greatest AUC when evaluating the different configurations of hyperparameters.

As a baseline, we evaluated a model based on the work by Hirano et al. [20] which classified mild liver fibrosis on NC CT images. This method passed the following three features extracted from 5 manually selected cubic ROIs to a logistic regression model with L2 regularization:

- 2D wavelet decomposition feature
- Standard deviation of variance filter
- Mean CT intensity

# Results

The patient demographics are summarized in Table 1.

Table 1: Demographics of study sample.

| Demographic | Statistic |
| --- | --- |
| Number of Patients | 169 |
| Mean Age ± Standard Deviation | 51.19 ± 14.70 |
| Number of Men | 101 |
| Number of Women | 68 |

## Performance of configurations

The models were evaluated based on their AUC. The mean test AUC on the biopsy-based ROIs and the non-biopsy ROIs for the considered settings are presented in Table 2, with the value resulting with the greatest mean AUC for each setting in bold. These results were evaluated using the internal test cohorts.

Table 2: Average test performance for all values from each setting.

| Setting | Value | Biopsy-based ROIs | Non-biopsy ROIs |
|---|---|---|---|
| Contrast | NC | **0.6658; 95% CI: [0.6484, 0.6833]** | **0.6100; 95% CI: [0.5897, 0.6303]** |
| | CE | 0.6115; 95% CI: [0.5915, 0.6315] | 0.5680; 95% CI: [0.5471, 0.5890] |
| Normalization | None | 0.6071; 95% CI: [0.5881, 0.6261] | 0.5919; 95% CI: [0.5714, 0.6123] |
| | Histogram Equalization | 0.6312; 95% CI: [0.6125, 0.6499] | 0.5488; 95% CI: [0.5283, 0.5693] |
| | Min-Max | 0.6532; 95% CI: [0.6345, 0.6718] | 0.5951; 95% CI: [0.5740, 0.6163] |
| | Z-Score | 0.6388; 95% CI: [0.6206, 0.6570] | 0.5938; 95% CI: [0.5731, 0.6145] |
| | Gamma-0.5 | 0.6292; 95% CI: [0.6100, 0.6484] | **0.6046; 95% CI: [0.5836, 0.6255]** |
| | Gamma-1.5 | **0.6726; 95% CI: [0.6539, 0.6913]** | 0.6000; 95% CI: [0.5800, 0.6200] |
| Model | Logistic Regression | **0.6558; 95% CI: [0.6375, 0.6741]** | **0.6054; 95% CI: [0.5853, 0.6256]** |
| | Random Forest | 0.6292; 95% CI: [0.6106, 0.6478] | 0.5817; 95% CI: [0.5611, 0.6023] |
| | SVM | 0.6329; 95% CI: [0.6136, 0.6522] | 0.5862; 95% CI: [0.5655, 0.6069] |
| | Linear with SGD | 0.6368; 95% CI: [0.6180, 0.6556] | 0.5827; 95% CI: [0.5617, 0.6038] |
| Feature Selection | None | 0.6286; 95% CI: [0.6103, 0.6469] | 0.5797; 95% CI: [0.5593, 0.6001] |
| | PCA | 0.6115; 95% CI: [0.5924, 0.6307] | 0.5744; 95% CI: [0.5533, 0.5955] |
| | Boruta | **0.6738; 95% CI: [0.6558, 0.6926]** | **0.6167; 95% CI: [0.5964, 0.6376]** |
| | LASSO Regression | 0.6408; 95% CI: [0.6221, 0.6594] | 0.58580; 95% CI: [0.5645, 0.6055] |

The five combinations of settings with the greatest AUCs on the internal test cohort using the biopsy-based ROIs and the non-biopsy ROIs presented in Table 3. Gamma-1.5 normalization and Boruta feature selection on non-contrast CT had the greatest mean AUC for their respective settings and were used in the top models evaluated on both sets of ROIs. The top model on biopsy-based ROIs had an AUC of 0.7390 and the top model on non-biopsy ROIs had an AUC of 0.6726. Logistic regression outperformed the other models, achieving test AUCs of 0.6558 and 0.6054 on biopsy-based and non-biopsy ROIs. Thus Gamma-1.5 normalization, Boruta feature selection, logistic regression on non-contrast CT images were selected as the hyperparameters to be used to train the simple models.

Table 3: Settings and AUC scores for the five best configurations evaluated on internal test cohorts of biopsy-based and non-biopsy ROIs.

| Test ROI | Settings | | | | AUC |
|---|---|---|---|---|---|
| | Normalization | Feature Selection | Model | Contrast | |
| Biopsy-based | Gamma-1.5 | Boruta | SGD | NC | 0.7390; 95% CI: [0.7229, 0.7551] |
| | Gamma-1.5 | Boruta | Logistic Regression | NC | 0.7356; 95% CI: [0.7188, 0.7524] |
| | Min-Max | Boruta | SGD | NC | 0.7329; 95% CI: [0.7153, 0.7504] |
| | Gamma-1.5 | LASSO | Logistic Regression | NC | 0.7302; 95% CI: [0.7135, 0.7470] |
| | Min-Max | Boruta | Logistic Regression | NC | 0.7256; 95% CI: [0.7071, 0.7440] |
| Non-biopsy | Gamma-1.5 | Boruta | Logistic Regression | NC | 0.6726; 95% CI: [0.6545, 0.6906] |
| | None | Boruta | SGD | NC | 0.6704; 95% CI: [0.6515, 0.6892] |
| | None | Boruta | Logistic Regression | NC | 0.6701; 95% CI: [0.6505, 0.6896] |
| | Gamma-1.5 | Boruta | SGD | NC | 0.6660; 95% CI: [0.6483, 0.6838] |
| | Gamma-1.5 | Boruta | SVM | NC | 0.6592; 95% CI: [0.6404, 0.6780] |

**Feature selection**

The 12 most common features across the top 5 models evaluated on biopsy-based ROIs and across the top five models evaluated on non-biopsy ROIs were ranked based on importance. The features were ranked by first filtering the top 5 features from each experiment and then counting the total number of occurrences of each filtered feature across all models. The top 12 features across the five best models evaluated on features from biopsy-based and non-biopsy ROIs can be found in Figure 4 and Figure 5, respectively. Feature rankings using all models including those outside of the 5 best configurations using each set of ROIs can be found in the supplementary material.

In the feature rankings for 5 best configurations from both the biopsy-based and non-biopsy ROIs presented in Figure 4 and Figure 5, the top 5 features were significantly more common than the other 7 features.

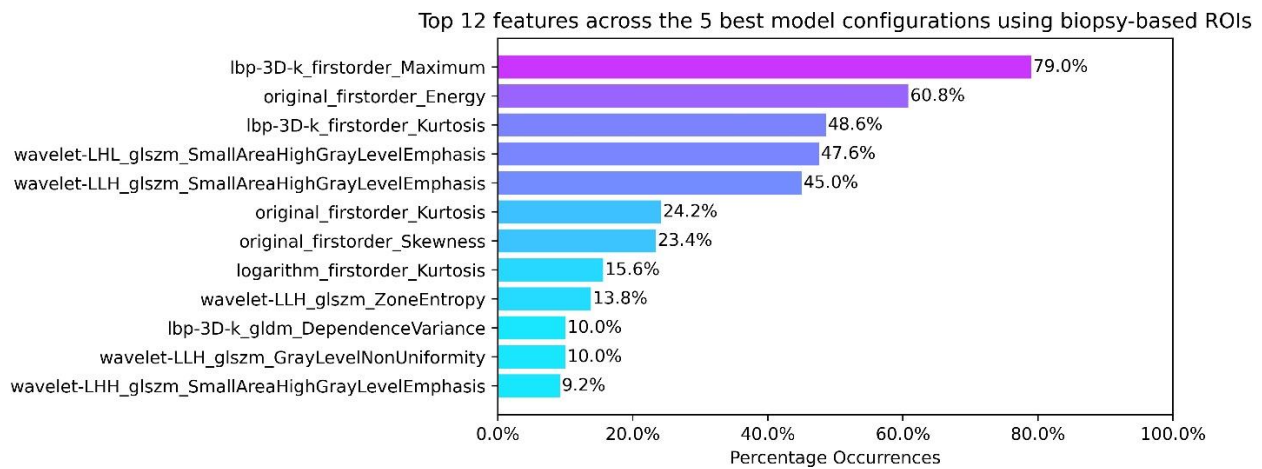

Figure 4: Top 12 features across the 5 best configurations using biopsy-based ROIs.

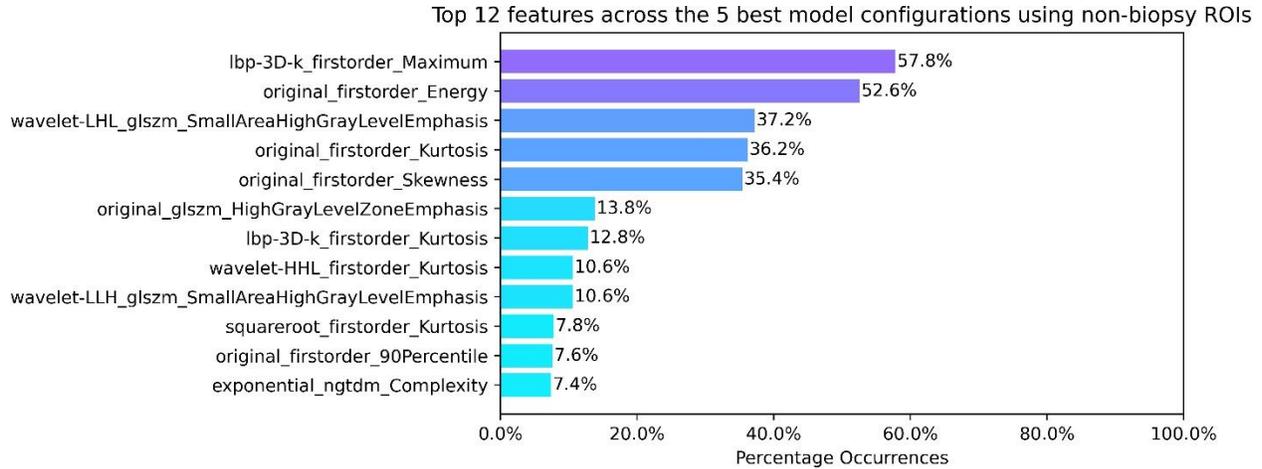

Figure 5: Top 12 features across the 5 best configurations using non-biopsy ROIs.

## Model performance using curated feature sets

Two sets of features arise as the most prominent from the feature rankings. The first is the top 5 features across the 5 best configurations using biopsy-based ROIs, which will be referred to as the biopsy-based feature set. These features are:

- lbp-3D-k_firstorder_Maximum
- original_firstorder_Energy
- lbp-3D-k_firstorder_Kurtosis
- wavelet-LHL_glszm_SmallAreaHighGrayLevelEmphasis
- wavelet-LLH_glszm_SmallAreaHighGrayLevelEmphasis

The second is the top 5 features across the 5 best configurations using non-biopsy ROIs, which will be referred to as the non-biopsy feature set. These features are:

- lbp-3D-k_firstorder_Maximum
- original_firstorder_Energy
- wavelet-LHL_glszm_SmallAreaHighGrayLevelEmphasis
- original_firstorder_Kurtosis
- original_firstorder_Skewness

Each set of features was used to train a simple model for the purpose of determining each feature set's effectiveness for fibrosis detection. These simple models were logistic regression models trained on NC CT images normalized using Gamma-1.5 normalization using Boruta feature selection. We chose these settings as they were the best-performing settings. These models were trained using the internal cohort and evaluated on the external held-out test cohort.

Table 4 presents the AUCs of the models on the external held-out test cohort, and Table 5 presents the sensitivity and specificity metrics for these models. The models presented in this table include the simple models trained on each feature set, the best configurations for the biopsy-based ROIs and the non-biopsy ROIs using all features, and a baseline model. The model trained using the non-biopsy feature set demonstrated overall higher AUC values (biopsy-based

ROIs: 0.8041, non-biopsy ROIs: 0.7833) as well as higher sensitivities (biopsy-based ROIs: 0.7227, non-biopsy ROIs: 0.9091).

Table 4: AUC of simple models compared to best models trained using all features, evaluated on both sets of ROIs from the external test cohort.

| Model | Biopsy-based ROIs | Non-biopsy ROIs |
|---|---|---|
| Simple Model Using Biopsy-based Feature Set | 0.7685; 95% CI: [0.7677, 0.7693] | 0.7905; 95% CI: [0.7893, 0.7916] |
| Simple Model Using Non-biopsy Feature Set | 0.8041; 95% CI: [0.8032, 0.8049] | 0.7833; 95% CI: [0.7821, 0.7845] |
| Best Model on Biopsy-based Internal Cohort Test ROIs Using All Features | 0.7397; 95% CI: [0.7357, 0.7437] | 0.7506; 95% CI: [0.7461, 0.7551] |
| Best Model on Non-biopsy Internal Cohort Test ROIs Using All Features | 0.7319; 95% CI: [0.7282, 0.7357] | 0.7447; 95% CI: [0.7405, 0.7488] |
| Baseline Model | 0.7843; 95% CI: [0.7824, 0.7862] | 0.7482; 95% CI: [0.7457, 0.7508] |

Table 5: Sensitivity and specificity of simple models compared to best models trained using all features, evaluated on both sets of ROIs from the external test cohort.

| Model | Biopsy-based ROIs | | Non-biopsy ROIs | |
|---|---|---|---|---|
| | Sensitivity | Specificity | Sensitivity | Specificity |
| Simple Model Using Biopsy-based Feature Set | 0.6446; 95% CI: [0.6378, 0.6513] | 0.6144; 95% CI: [0.6091, 0.6196] | 0.8936; 95% CI: [0.8869, 0.9003] | 0.5661; 95% CI: [0.5649, 0.5673] |
| Simple Model Using Non-biopsy Feature Set | 0.7227; 95% CI: [0.7181, 0.7274] | 0.5217; 95% CI: [0.5217, 0.5217] | 0.9091; 95% CI: [0.9091, 0.9091] | 0.5283; 95% CI: [0.5231, 0.5334] |
| Best Model on Biopsy-based Internal Cohort Test ROIs Using All Features | 0.7791; 95% CI: [0.7380, 0.8202] | 0.5630; 95% CI: [0.5348, 0.5913] | 0.8655; 95% CI: [0.8202, 0.9107] | 0.5574; 95% CI: [0.5261, 0.5887] |
| Best Model on Non-biopsy Internal Cohort Test ROIs Using All Features | 0.8118; 95% CI: [0.8038, 0.8199] | 0.5209; 95% CI: [0.5120, 0.5297] | 0.9718; 95% CI: [0.9609, 0.9827] | 0.5322; 95% CI: [0.5278, 0.5365] |
| Baseline Model | 0.9400; 95% CI: [0.9316, 0.9484] | 0.6035; 95% CI: [0.6007, 0.60625] | 1.000; 95% CI: [1.000, 1.000] | 0.5657; 95% CI: [0.5648, 0.5665] |

## Discussion

Our study of ML-based liver fibrosis detection on CT images using radiomic features demonstrated that using NC CT images normalized using Gamma-1.5 was the overall most effective approach for fibrosis detection. This indicates that despite humans identifying fibrosis more easily in CE CT images, the CE CT images obfuscate valuable information that is present on NC CT images. We also demonstrated that for these images, logistic regression classifiers using feature inputs of lbp-3D-k_firstorder_Maximum, original_firstorder_Energy, original_firstorder_Kurtosis, original_firstorder_Skewness, and wavelet-LHL_glszm_SmallAreaHighGrayLevelEmphasis are the most effective for liver fibrosis detection. We trained a model using these insights to produce a highly effective classifier for liver fibrosis detection on CT images. We evaluated the model on random liver ROIs as indicate the model's performance as screening tools when biopsy locations are not available.

      To contextualize our work, Lubner et al. achieved an AUC of 0.78 using the mean gray-level intensity feature to detect liver fibrosis on CT images [21]. Hirano et al. also aimed to detect liver fibrosis on CT images by proposing a model using features consisting of a 2D wavelet decomposition feature, standard deviation of variance filter, and mean CT intensity [20]. Our model achieved AUCs of 0.8041 and 0.7833 on biopsy-based and non-biopsy ROIs respectively, which are higher than those reported by Lubner et al. in their work, and outperformed the model proposed by Hirano et al. when evaluated on our dataset. For other imaging modalities, House et al. achieved an AUC of 0.91 using MRI [22] while Zhang et al. found that MRI is more effective for liver fibrosis staging [23], indicating that MRI is superior to CT purely for liver fibrosis detection. However, given the limited access to MRI compared to CT, the increased costs, and the need for dedicated acquisitions to acquire the images, the value and the use cases of CT and MRI can be considered distinct from one another.

      One notable limitation of our study is the selection of ROIs. We used ROIs distal to the biopsy needle due to concern that designating ROIs significantly distant from the biopsy needle might result in an ROI that captured an area with a different fibrosis stage than the fibrosis stage specified from biopsy due to the geographic distribution of the disease. While this issue is addressed by also using random non-biopsy ROIs, the use of biopsy-based ROIs in a study on non-invasive liver fibrosis detection can introduce unexpected bias to the model, despite the benefits that come with using ROIs spatially and temporally synchronized with the ground truth biopsy measurements. Furthermore, the ROIs were placed based on a single human reader, introducing a risk that another human reader would have placed slightly different ROIs thereby impacting the radiomic features. Hirano et al. achieved an AUC value of 0.86 in their work [20], but when applying their method as a baseline model, the baseline model achieved an AUC of 0.7843. The major difference between the original implementation and our implementation was the ROI selection, indicating that ROI selection can have significant effects on the observed AUC. Another limitation of this work is that while it was found that Gamma correction with $\gamma = 1.5$ outperformed Gamma correction with $\gamma = 0.5$, higher $\gamma$ values were not explored due to computational cost. It is possible that higher values could further improve classification

performance. In addition, our proposed model cannot be used in cases where only CE images are available due to it being trained on NC images.

In conclusion, we present a model for use as a non-invasive liver fibrosis screening tool on CT images acquired for any clinical reason. Our model achieved an AUC of 0.7833 and a sensitivity of 0.9091 on random ROIs, indicating that our proposed model is effective for the purposes of screening patients for liver fibrosis. The model can be applied to random liver ROIs extracted from CT images detection of liver fibrosis on NC CT images. Patients detected by the model to have liver fibrosis can then be recommended for evaluation using biopsy or MRI to better stage the fibrosis and begin developing a treatment plan. Our work enables this process, thereby improving access to liver fibrosis detection and increasing the number of patients identified to have early subclinical stages, when the disease progression can still be acted upon.

# Supplementary Material

## Model specific hyperparameters evaluated in grid-search

The machine learning (ML) model specific settings which differed from the primary settings explored in the main manuscript and were evaluated during the grid-search are based on their implementations in the scikit-learn Python library and are presented in Table 1.

Table 1: Model type-specific settings and their candidate values.

| Model Type | Hyperparameter | Candidate Values |
|---|---|---|
| Logistic Regression | Optimization Solver | L-BFGS-B |
| | | Newton-CG |
| | | Stochastic Average Gradient (SAG) |
| | | Stochastic Average Gradient Ascent (SAGA) |
| | Inverse Regularization Strength | 0.5 |
| | | 1.0 |
| | | 1.5 |
| Random Forest | Number of Estimators | 50 |
| | | 100 |
| | | 200 |
| | Maximum Number of Features | Auto |
| | | Square Root |
| | Maximum Depth | All |
| | | 5 |
| | | 100 |
| SVM | Kernel | Linear |
| | | Polynomial (degree 3) |
| | | Radial Basis Function (RBF) |
| | Inverse L2 Squared Regularization Strength | 0.5 |
| | | 1.0 |
| | | 1.5 |
| | | 0.001 |
| Linear Model with SGD | L2 Regularization Strength | 0.0001 |
| | | 0.00001 |
| | Loss Function | Logistic Regression |
| | | Modified Huber Loss |

**Feature rankings**

Feature rankings of the top 12 features across all models on both biopsy-based regions of interest (ROIs) and non-biopsy ROIs were calculated. The feature rankings corresponding to biopsy-based ROIs and non-biopsy ROIs are visualized in Figure 1 and Figure 2 respectively.

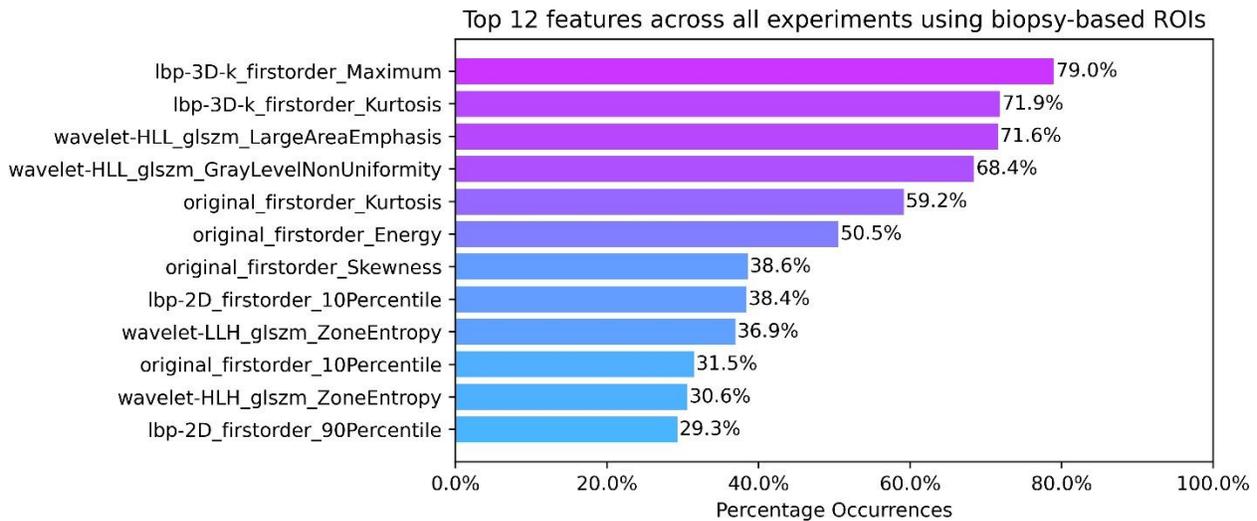

Figure 1: Top 12 features across all experiments using biopsy-based ROIs.

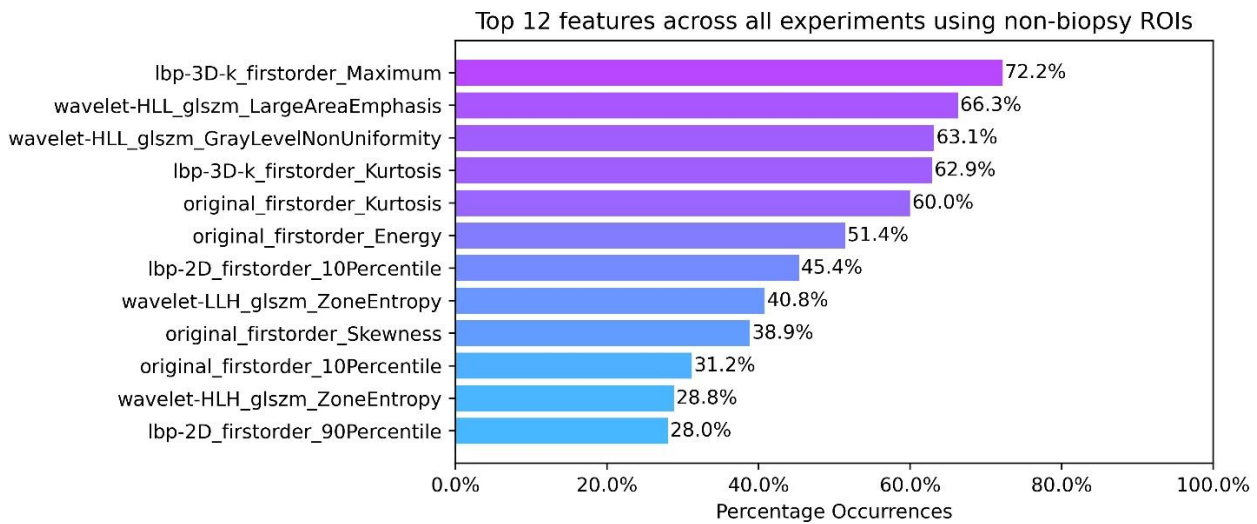

Figure 2: Top 12 features across all experiments using non-biopsy ROIs.

Based on the feature rankings across all experiments, one additional set of features was considered. This set is the set of features that appeared in the top 12 features across all experiments and across the top 5 models for both sets of ROIs. In simpler terms, the features that appeared in Figure 1 and Figure 2 as well as the feature ranking figures from the main manuscript. These features are lbp-3D-k_firstorder_Kurtosis, lbp-3D-k_firstorder_Maximum, original_firstorder_Energy, original_firstorder_Kurtosis, original_firstorder_Skewness and will be referred to as the intersecting feature set.

Using these features to train logistic regression models trained on NC images normalized using Gamma correction with $\gamma = 1.5$ (Gamma-1.5) normalization using Boruta feature selection yielded the results presented in Table 2. This table includes the other models and baselines that were compared in the main manuscript.

Table 2: AUC of simple models trained on intersecting feature set compared to other models, evaluated on both sets of ROIs from the external test cohort.

| Model | Biopsy-based ROIs | Non-biopsy ROIs |
|---|---|---|
| Simple Model Using Intersecting Feature Set | 0.7536; 95% CI: [0.7528, 0.7544] | 0.7464; 95% CI: [0.7451, 0.7476] |
| Simple Model Using Biopsy-based Feature Set | 0.7685; 95% CI: [0.7677, 0.7693] | 0.7905; 95% CI: [0.7893, 0.7916] |
| Simple Model Using Non-biopsy Feature Set | 0.8041; 95% CI: [0.8032, 0.8049] | 0.7833; 95% CI: [0.7821, 0.7845] |
| Best Model on Biopsy-based Internal Cohort Test ROIs Using All Features | 0.7397; 95% CI: [0.7357, 0.7437] | 0.7506; 95% CI: [0.7461, 0.7551] |
| Best Model on Non-biopsy Internal Cohort Test ROIs Using All Features | 0.7319; 95% CI: [0.7282, 0.7357] | 0.7447; 95% CI: [0.7405, 0.7488] |
| Baseline Model | 0.7843; 95% CI: [0.7824, 0.7862] | 0.7482; 95% CI: [0.7457, 0.7508] |

## Robustness to region of interest volume confoundment

Welch et al. [24] found that features from PyRadiomics that are dependent on image attenuation such as energy can effectively serve as surrogates for delineated tumor volume, even when shape features such as volume are not meant to be considered. To ensure that the findings of this work are independent of the ROI volume, a model was trained using the MeshVolume feature, the primary volume metric for PyRadiomics, and another model was trained using the pixel spacing values of each image. Both models were trained using NC images normalized using Gamma-1.5 normalization as was done for the simple model in the Model Performance Using Curated Feature Sets Section of the main document. No feature selection was used to ensure that the volume-related features were considered by the models. The models trained using MeshVolume were trained using a logistic regression model while the model trained using pixel spacing values were trained using logistic regression models and random forest models. The pixel spacing serves as an effective measure of ROI volume as all ROIs are spheres with 1.5 cm radii. Thus, the differences between the volumes of ROIs between images can be measured using the pixel spacing as they impact how many voxels constitute the ROI.

The performances of these models are presented in Table 3. These results indicate that while there is some volume information, there is significant additional signal present beyond volume in the models trained using PyRadiomics features in this work. Thus, the volume information is ultimately not a strong confounding factor in the results presented in the main document.

Table 3: Mean test AUC of volume dependent models.

| Model | Biopsy-based ROIs | Non-biopsy ROIs |
|---|---|---|
| Mesh Volume Models | 0.4980; 95% CI: [0.4980, 0.4980] | 0.5968; 95% CI: [0.5968, 0.5968] |
| Pixel Spacing Models | 0.5524; 95% CI: [0.5508, 0.5540] | 0.5524; 95% CI: [0.5508, 0.5540] |

# Funding

This work was supported by Chair in Medical Imaging and Artificial Intelligence, a joint Hospital-University Chair between the University of Toronto, The Hospital for Sick Children, and the SickKids Foundation.